\documentclass{article}
\usepackage[preprint]{spconf}
\usepackage{amsmath,graphicx}
\usepackage{amssymb}
\usepackage{subcaption}
\usepackage[hidelinks]{hyperref}
\usepackage{url}

\usepackage{tikz}
\usetikzlibrary{shapes, arrows, positioning, decorations.pathreplacing, calc}

\graphicspath{{./figs/}}

\newcommand{\twodots}{\mathinner {\ldotp \ldotp}}

\title{Learning Convolutional Transforms for Lossy Point Cloud Geometry Compression}
\name{Maurice Quach \qquad Giuseppe Valenzise \qquad Frederic Dufaux}
\address{L2S, CNRS, CentraleSup\'elec, Universit\'e Paris-Saclay}
\copyrightnotice{ICIP 2019}
\toappear{Copyright 2019 IEEE}

\begin{document}
\maketitle
\begin{abstract}
Efficient point cloud compression is fundamental to enable the deployment of virtual and mixed reality applications, since the number of points to code can range in the order of millions.
In this paper, we present a novel data-driven geometry compression method for static point clouds based on learned convolutional transforms and uniform quantization.
We perform joint optimization of both rate and distortion using a trade-off parameter. In addition, we cast the decoding process as a binary classification of the point cloud occupancy map.
Our method outperforms the MPEG reference solution in terms of rate-distortion on the Microsoft Voxelized Upper Bodies dataset with $51.5\%$ BDBR savings on average.
Moreover, while octree-based methods face exponential diminution of the number of points at low bitrates, our method still produces high resolution outputs even at low bitrates.
Code and supplementary material are available at \url{https://github.com/mauriceqch/pcc_geo_cnn}.
\end{abstract}


\section{Introduction}
\label{sec:intro}
Point clouds are an essential data structure for Virtual Reality (VR) and Mixed Reality (MR) applications.
A point cloud is a set of points in the 3D space represented by coordinates $x, y, z$ and optional attributes (for example color, normals, etc.).
Point cloud data is often very large as point clouds easily range in the millions of points and can have complex sets of attributes.
Therefore, efficient point cloud compression (PCC) is particularly important to enable practical usage in VR and MR applications.

The Moving Picture Experts Group (MPEG) is currently working on PCC.
In 2017, MPEG issued a call for proposals (CfP) and in order to provide a baseline, a point cloud codec for tele-immersive video \cite{mekuria_design_2017} was chosen as the MPEG anchor.
To compare the proposed compression solutions, quality evaluation metrics were developed leading to the selection of the point-to-point (D1) and point-to-plane (D2) as baseline metrics \cite{schwarz_common_2018}.
The point to point metric, also called D1 metric, is computed using the Mean Squared Error (MSE) between the reconstructed points and the nearest neighbors in the reference point cloud.
The point-to-plane metric, also called D2 metric, uses the surface plane instead of the nearest neighbor.

Research on PCC can be categorized along two dimensions.
On one hand, one can either compress point cloud geometry, i.e., the spatial position of the points, or their associated attributes.
On the other hand, we can also separate works focusing on compression of dynamic point clouds, which contain temporal information, and static point clouds.

In this work, we focus on the lossy compression of static point cloud geometry.
In PCC, a precise reconstruction of geometric information is of paramount importance to enable high-quality rendering and interactive applications. For this reasons, lossless geometry coding has been investigated recently in MPEG, but even state-of-the-art techniques struggle to compress beyond about 2 bits per occupied voxels (bpov)~\cite{garcia_intra-frame_2018}. This results in large storage and transmission costs for rich point clouds.
Lossy compression proposed in the literature, on the other hand, are based on octrees which achieve variable-rate geometry compression by changing the octree depth.
Unfortunately, lowering the depth reduces the number of points exponentially.
As a result, octree based lossy compression tends to produce ``blocky'' results at the rendering stage with medium to low bitrates.
In order to partially attenuate this issue, \cite{krivokuca_volumetric_2018} proposes to use wavelet transforms and volumetric functions to compact the energy of the point cloud signal. However, since they still employ an octree representation, their method exhibits rapid geometry degradation at lower bitrates.
While previous approaches use hand-crafted transforms, we propose here a data driven approach based on learned convolutional transforms which directly works on voxels.

Specifically, we present a method for learning analysis and synthesis transforms suitable for point cloud geometry compression. In addition, by interpreting the point cloud geometry as a binary signal defined over the voxel grid, we cast decoding as the problem of classifying whether a given voxel is occupied or not.
We train our model on the ModelNet40 mesh dataset \cite{wu_3d_2015, sedaghat_orientation-boosted_2016}, test its performance on the Microsoft Voxelized Upper Bodies (MVUB) dataset \cite{loop_microsoft_2016} and compare it with the MPEG anchor \cite{mekuria_design_2017}.
We find that our method outperforms the anchor on all sequences at all bitrates.
Additionally, in contrast to octree-based methods, ours does not exhibit exponential diminution in the number of points when lowering the bitrate.
We also show that our model generalizes well by using completely different datasets for training and testing.

After reviewing related work in Section~\ref{sec:related}, we describe the proposed method in Section~\ref{sec:proposed} and evaluate it on different datasets in Section~\ref{sec:experimental}. Conclusions are drawn in Section~\ref{sec:conclusion}.

\section{Related Work}
\label{sec:related}
Our work is mainly related to point cloud geometry compression, deep learning based  image and video compression and applications of deep learning to 3D objects.

Point cloud geometry compression research has mainly focused on tree based methods \cite{garcia_intra-frame_2018, krivokuca_volumetric_2018, mekuria_design_2017} and dynamic point clouds \cite{queiroz_distance-based_2018, thanou_graph-based_2016}.
Our work takes a different approach by compressing point cloud geometry using a 3D auto-encoder.
While classical compression approaches use hand-crafted transforms, we directly learn the filters from data.

Recent research has also applied deep learning to image and video compression.
In particular, auto-encoders, recurrent neural networks and context-based learning have been used for image and video compression \cite{valenzise_quality_2018, balle_variational_2018, wang_enhancing_2019}.
\cite{balle_end--end_2017} proposes to replace quantization with additive uniform noise during training while performing actual quantization during evaluation.
Our work takes inspiration from this approach in the formulation of quantization, but significantly expands it with new tools, a different loss function and several practical adaptations to the case of point cloud geometry compression.

Generative models \cite{achlioptas_learning_2018} and auto-encoders \cite{girdhar_learning_2016} have also been employed to learn a latent space of 3D objects.
In the context of point cloud compression, our work differs from the above-mentioned approaches in two aspects.
First, we consider quantization in the training in order to jointly optimize for rate-distortion (RD) performance; second, we propose a lightweight architecture which allows us to process voxels grids with resolutions that are an order of magnitude higher than previous art.

\section{Proposed method}
\label{sec:proposed}

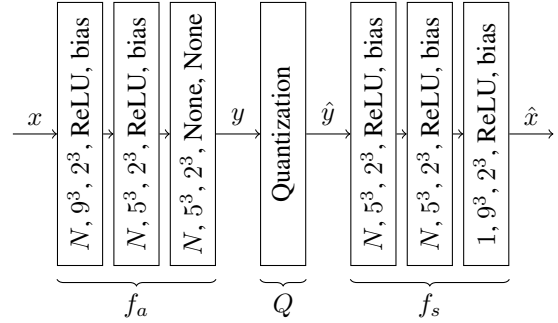
\begin{figure}
\centering
\begin{tikzpicture}[
		mainnode/.style={draw, rectangle, minimum width=3.5cm, minimum height=0.6cm, rotate=90}
	]
	\node (in) [rotate=90] {};
	\node (box1) [mainnode, right=1.20cm of in.south, anchor=south] {$N$, $9^{3}$, $2^{3}$, ReLU, bias};
	\node (box2) [mainnode, right=0.75cm of box1.south, anchor=south] {$N$, $5^{3}$, $2^{3}$, ReLU, bias};
	\node (box3) [mainnode, right=0.75cm of box2.south, anchor=south] {$N$, $5^{3}$, $2^{3}$, None, None};

	\node (quant) [mainnode, right=1.20cm of box3.south, anchor=south] {Quantization};

	\node (box4) [mainnode, right=1.20cm of quant.south, anchor=south] {$N$, $5^{3}$, $2^{3}$, ReLU, bias};
	\node (box5) [mainnode, right=0.75cm of box4.south, anchor=south] {$N$, $5^{3}$, $2^{3}$, ReLU, bias};
	\node (box6) [mainnode, right=0.75cm of box5.south, anchor=south] {$1$, $9^{3}$, $2^{3}$, ReLU, bias};

	\node (out) [right=1.20cm of box6.south, anchor=south, rotate=90, minimum height=0.6cm] {};

	\draw [->] (in) -- node [above] {$x$} (box1);
	\draw [->] (box1) -- (box2);
	\draw [->] (box2) -- (box3);
	\draw [->] (box3) -- node [above] {$y$} (quant);
	\draw [->] (quant) -- node [above] {$\hat{y}$} (box4);
	\draw [->] (box4) -- (box5);
	\draw [->] (box5) -- (box6);
	\draw [->] (box6) -- node [above] {$\hat{x}$} (out);

	\draw[decorate, decoration={brace}] let \p1=(box3.south west), \p2=(box1.north west) in ($(\x1, \y1-0.5em)$) -- ($(\x2, \y2-0.5em)$) node [below, midway, below=2pt] {$f_{a}$};
	\draw[decorate, decoration={brace}] let \p1=(quant.south west), \p2=(quant.north west) in ($(\x1, \y1-0.5em)$) -- ($(\x2, \y2-0.5em)$) node [below, midway, below=2pt] {$Q$};
	\draw[decorate, decoration={brace}] let \p1=(box6.south west), \p2=(box4.north west) in ($(\x1, \y1-0.5em)$) -- ($(\x2, \y2-0.5em)$) node [below, midway, below=2pt] {$f_{s}$};
\end{tikzpicture}
\caption{Neural Network Architecture. Layers are specified using the following format: number of feature maps, filter size, strides, activation and bias.}
\label{fig:arch}
\end{figure}

In this section, we describe the proposed method in more details.

\subsection{Definitions}

First, we define the set of possible points at resolution $r$ as $\Omega_{r} = [0 \twodots r]^{3}$.
Then, we define a point cloud as a set of points $S \subseteq \Omega_{r}$ and its corresponding voxel grid $v_{S}$ as follows:
\begin{align*}
	v_{S} \colon \Omega_{r} & \longrightarrow \{0, 1\}, \\
	z & \longmapsto \begin{cases}
			1,& \text{if } z \in S\\
			0,& \text{otherwise}.
		\end{cases}
\end{align*}

For notational convenience, we use $s^{3}$ instead of $s \times s \times s$ for filter sizes and strides.

\subsection{Model}

We use a 3D convolutional auto-encoder composed of an analysis transform $f_{a}$, followed by a uniform quantizer and a synthesis transform $f_{s}$.

Let $x = v_{S}$ be the original point cloud.
The corresponding latent representation is $y = f_{a}(x)$.
To quantize $y$, we introduce a quantization function $Q$ so that $\hat{y} = Q(y)$.
This allows us to express the decompressed point cloud as $\hat{x} = \hat{v}_{S} = f_{s}(\hat{y})$.
Finally, we obtain the decompressed point cloud $\tilde{x} = \tilde{v}_{S} = round(\min(0, \max(1, \hat{x})))$ using element-wise minimum, maximum and rounding functions.

\begin{figure}
\centering
\begin{subfigure}{0.45\textwidth}
	\centering
	\includegraphics[width=6.5cm]{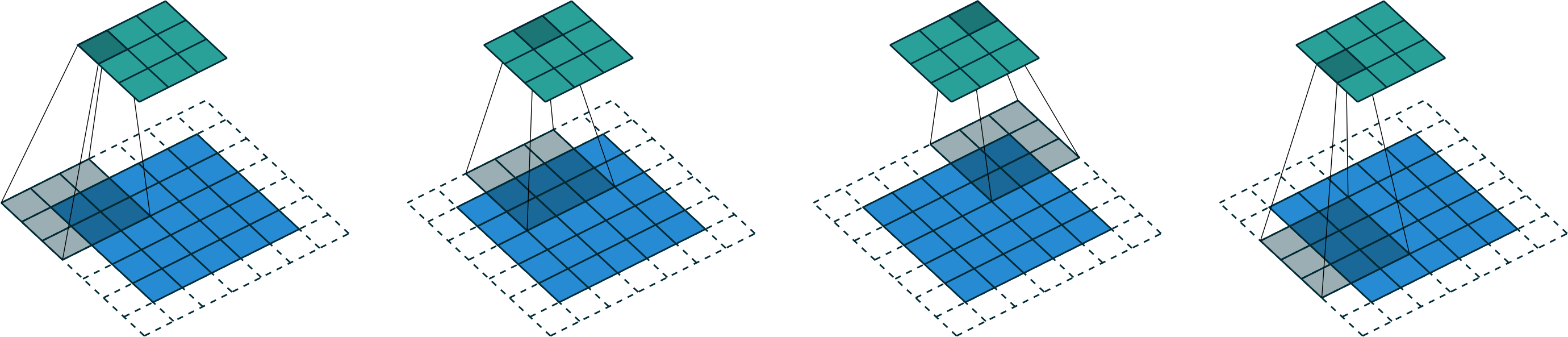}
	\caption{Strided convolution on a $5^{2}$ input with a $3^{2}$ filter, $2^{2}$ strides and same padding. The shape of the output is $3^{3}$.}
	\vspace{0.3cm}
\end{subfigure}
\begin{subfigure}{0.45\textwidth}
	\centering
	\includegraphics[width=6.5cm]{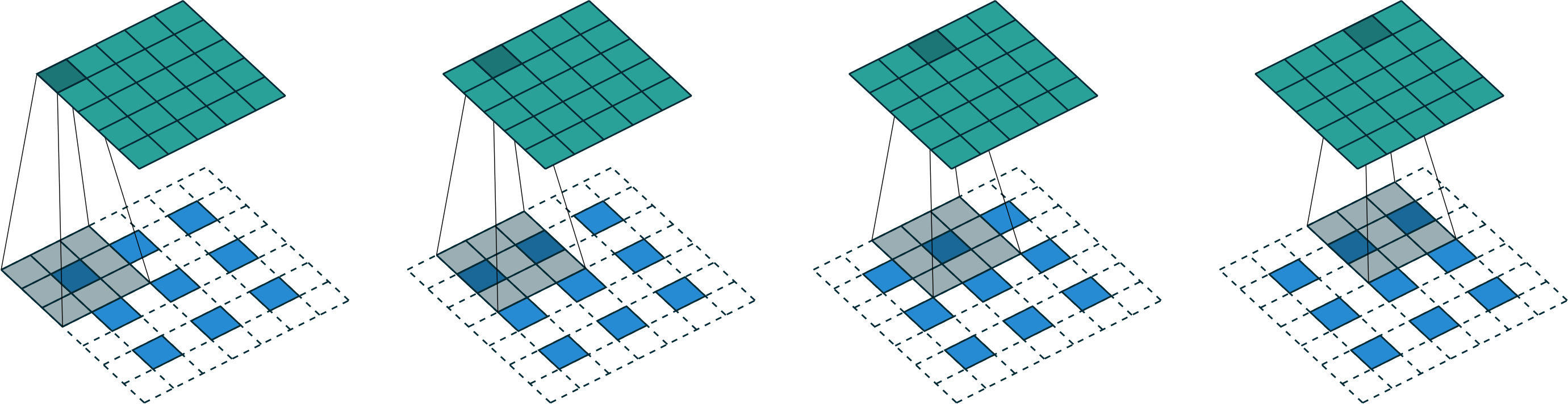}
	\caption{Strided transpose convolution on a $3^{2}$ input with a $3^{2}$ filter, $2^{2}$ strides and same padding. The shape of the output is $5^{2}$.}
\end{subfigure}
\caption{Strided convolution and strided transpose convolution operations. Illustrations from \cite{dumoulin_guide_2016}.}
\label{fig:conv}
\end{figure}

In our model, we use convolutions and transpose convolutions with same padding and strides.
They are illustrated in Figure \ref{fig:conv} and defined as follows :
\begin{itemize}
	\item Same (half) padding pads the input with zeros so that the output size is equal to the input size.
	\item Convolution performed with unit stride means that the convolution filter is computed for each element of the input array.
When iterating the input array, strides specify the step for each axis.
	\item Convolution can be seen as matrix multiplication and transpose convolution can be derived from this.
In particular, we can build a sparse matrix $C$ with non-zero elements corresponding to the weights.
The transpose convolution, also called \emph{deconvolution}, is obtained using the matrix $C^{T}$ as a layout for the weights.
\end{itemize}

Using these convolutional operations as a basis, we learn analysis and synthesis transforms structured as in Figure \ref{fig:arch} using the Adam optimizer \cite{kingma_adam:_2014} which is based on adaptive estimates of first and second moments of the gradient.

We handle quantization similarly to \cite{balle_variational_2018}.
$Q$ represents element-wise integer rounding during evaluation and $Q$ adds uniform noise between $-0.5$ and $0.5$ to each element during training which allows for differentiability.
To compress $Q(y)$, we perform range coding and use the Deflate algorithm, a combination of LZ77 and Huffman coding \cite{huffman_method_1952} with shape information on $x$ and $y$ added before compression.
Note however that our method does not assume any specific entropy coding algorithm.

Our decoding process can also be interpreted as a binary classification problem where each point $z \in \Omega_{r}$ of the voxel grid is either present or not.
This allows us to decompose $\hat{x} = \hat{v}_{S}$ into its individuals voxels $z$ whose associated value is $p_{z}$.
However, as point clouds are usually very sparse, most $v_{S}(z)$ values are equal to zero.
To compensate for the imbalance between empty and occupied voxels we use the $\alpha$-balanced focal loss as defined in \cite{lin_focal_2017}:
\begin{equation}
	FL(p_{z}^{t}) = -\alpha_{z}(1-p_{z}^{t})^{\gamma}log(p_{z}^{t})
\end{equation}
with $p_{z}^{t}$ defined as $p_{z}$ if $v_{S}(z) = 1$ and $1 - p_{z}$ otherwise.
Analogously, $\alpha_{z}$ is defined as $\alpha$ when $v_{S}(z) = 1$ and $1 - \alpha$ otherwise.
The focal loss for the decompressed point cloud can then be computed as follows:
\begin{equation}
	FL(\tilde{x}) = \sum_{z \in S} FL(p_{z}^{t}).
\end{equation}

Our final loss is $L = \lambda D + R$ where $D$ is the distortion calculated using the focal loss and $R$ is the rate in number of bits per input occupied voxel (bpov).
The rate is computed differently during training and during evaluation.
On one hand, during evaluation, as the data is quantized, we compute the rate using the number of bits of the final compressed representation.
On the other hand, during training, we add uniform noise in place of discretization to allow for differentiation.
It follows that the probability distribution of the latent space $Q(y)$ during training is a continuous relaxation of the probability distribution of $Q(y)$ during evaluation which is discrete.
As a result, entropies computed during training are actually differential entropies, or continuous entropies, while entropies computed during evaluation are discrete entropies.
During training, we use differential entropy as an approximation of discrete entropy.
This makes the loss differentiable which is primordial for training neural networks.

\section{Experimental results}
\label{sec:experimental}

We use train, evaluation and test split across two datasets.
We train and evaluate our network on the ModelNet40 aligned dataset \cite{wu_3d_2015, sedaghat_orientation-boosted_2016}.
Then, we perform tests on the MVUB dataset and we compare our method with the MPEG anchor \cite{mekuria_design_2017}.

We perform our experiments using Python 3.6 and Tensorflow 0.12. 
We use $N = 32$ filters, a batch size of $64$ and Adam with $lr = 10^{-4}$, $\beta_{1} = 0.9$ and $\beta_{2} = 0.999$.
For the focal loss, we use $\alpha = 0.9$ and $\gamma = 2.0$.

To compute distortion, we use the point-to-plane symmetric PSNR computed with the \emph{pc\_error} MPEG tool \cite{tian_geometric_2017}.

\subsection{Datasets}

The ModelNet40 dataset contains 12,311 mesh models from 40 categories.
This dataset provides us with both variety and quantity to ensure good generalization when training our network.
To convert this dataset to a point cloud dataset, we first perform sampling on the surface of each mesh. Then, we translate and scale it into a voxel grid of resolution $r$.
We use this dataset for training with a resolution $r = 64$.

The MVUB dataset \cite{loop_microsoft_2016} contains 5 sequences captured at 30 fps during 7 to 10 seconds each with a total of 1202 frames.
We test our method on each individual frame with a resolution $r = 512$.
In other words, we evaluate performance for intra-frame compression on each sequence.

\begin{figure*}[htb]
\centering
\begin{subfigure}{.33\textwidth}
\centering
\includegraphics[width=\linewidth]{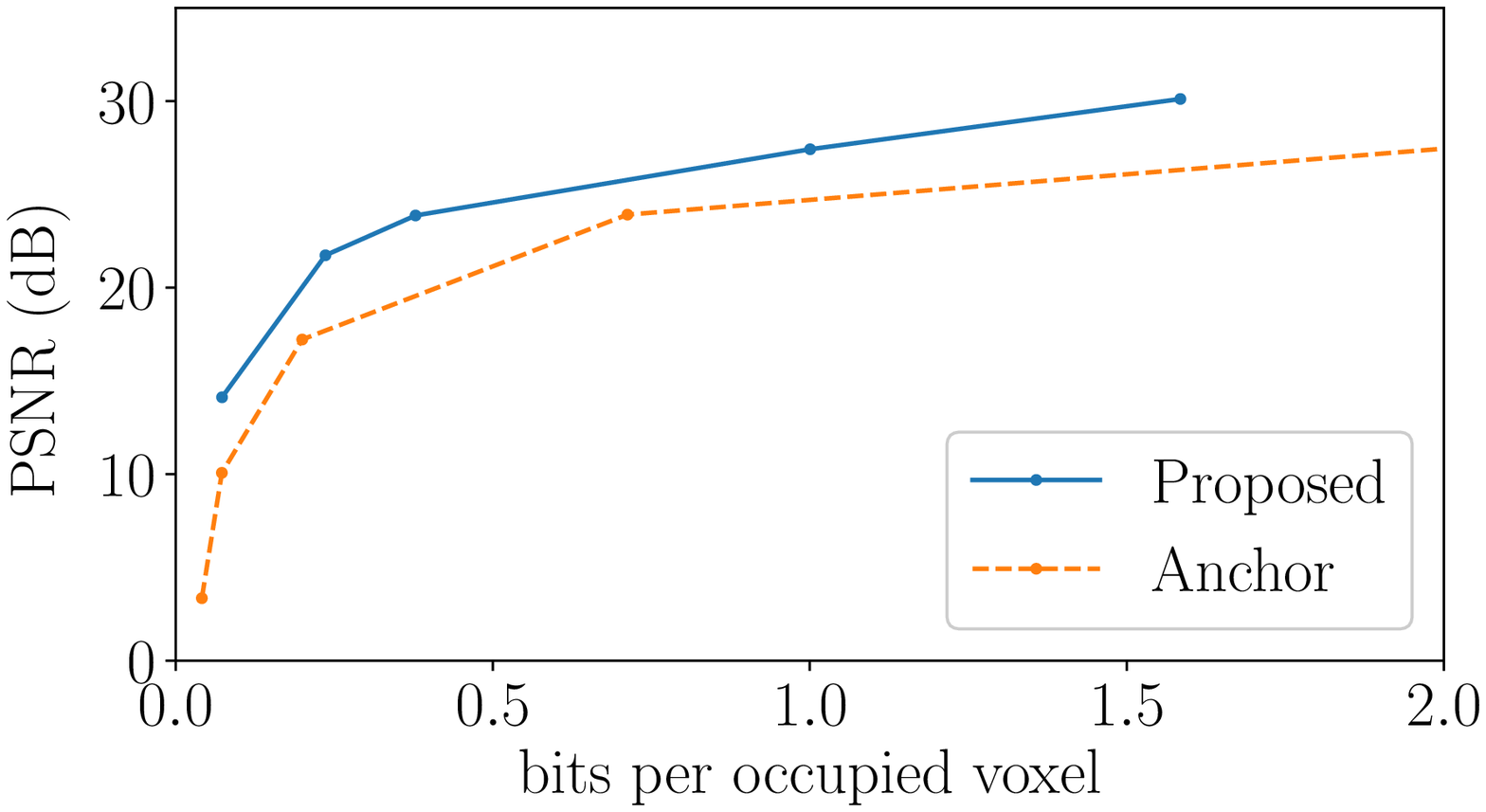}
\caption{Andrew sequence ($-47.8\%$ BDBR)}
\end{subfigure}
\hfill
\begin{subfigure}{.33\textwidth}
\centering
\includegraphics[width=\linewidth]{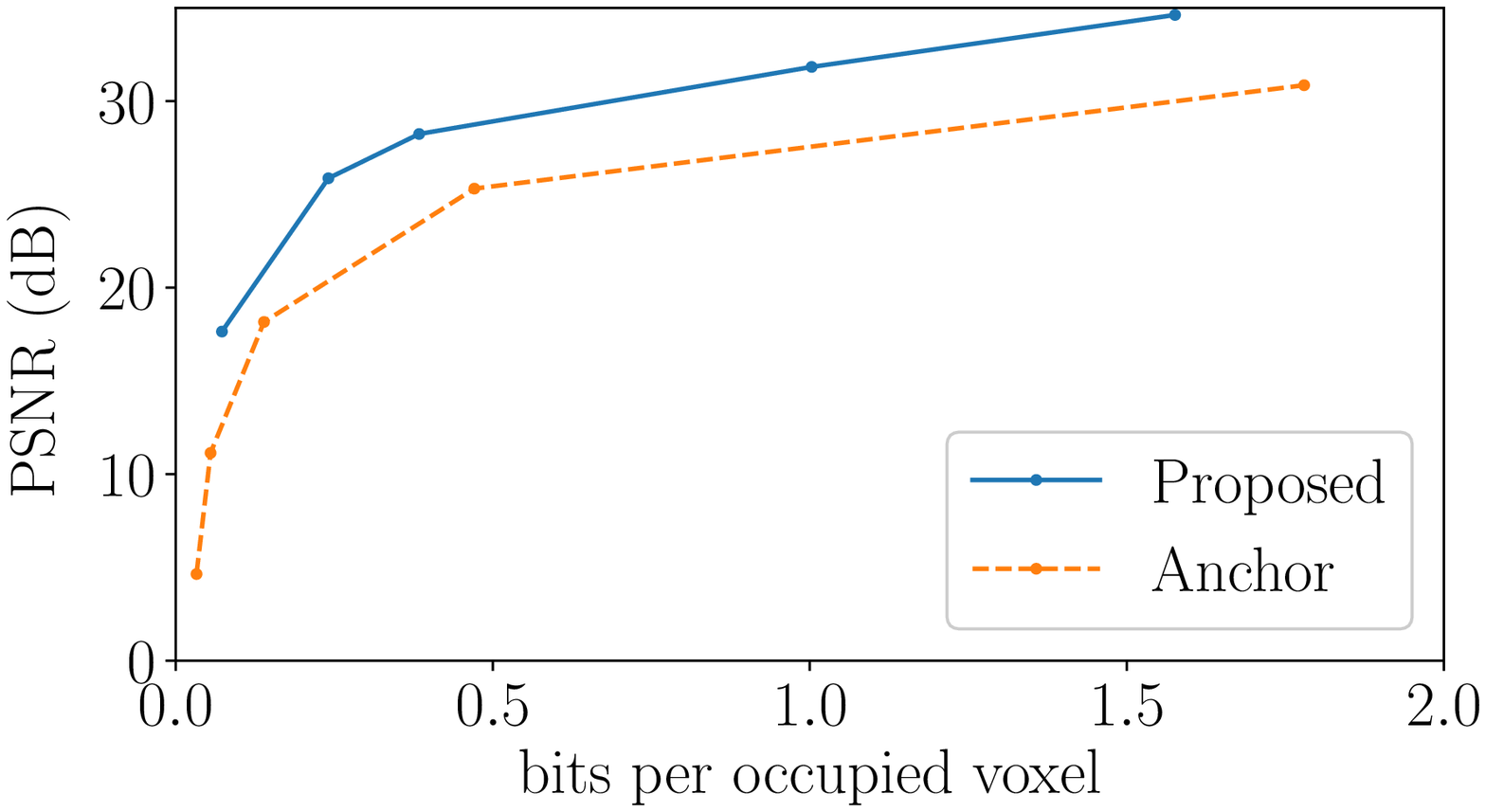}
\caption{David sequence ($-55.7\%$ BDBR)}
\end{subfigure}
\hfill
\begin{subfigure}{.33\textwidth}
\centering
\includegraphics[width=\linewidth]{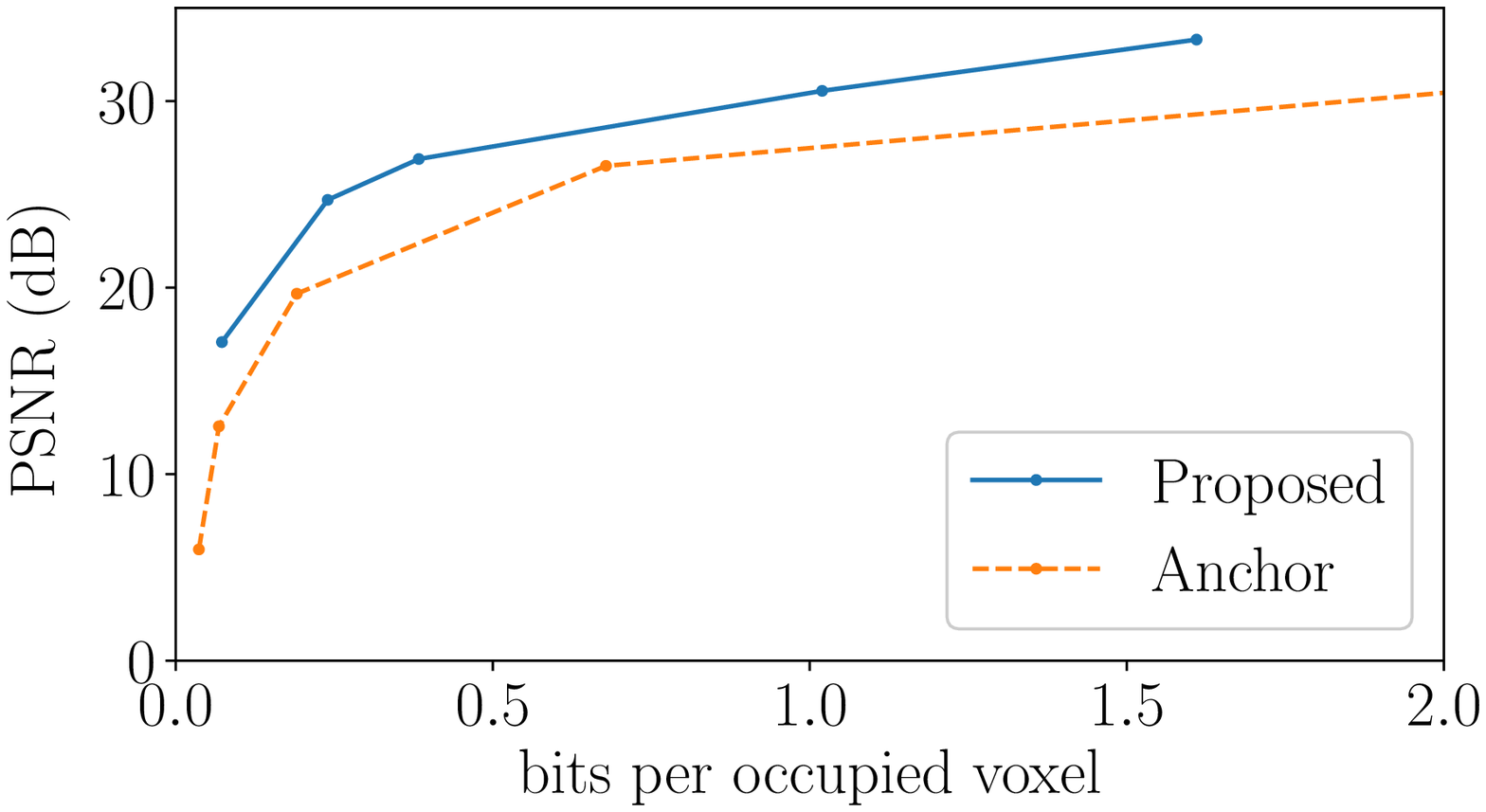}
\caption{Phil sequence ($-49.0\%$ BDBR)}
\end{subfigure}
\hfill
\begin{subfigure}{.33\textwidth}
\centering
\includegraphics[width=\linewidth]{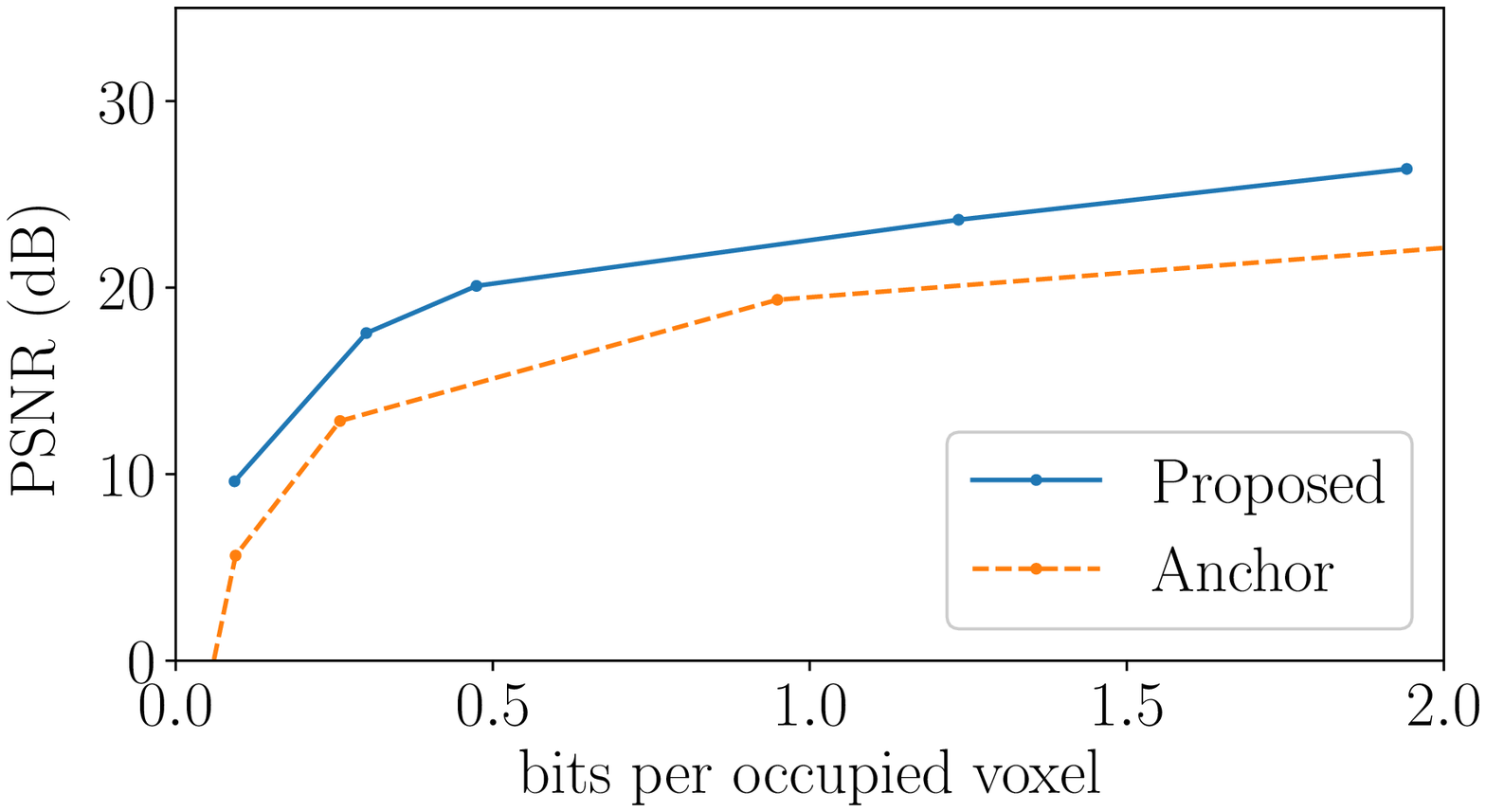}
\caption{Ricardo sequence ($-52.7\%$ BDBR)}
\end{subfigure}
\begin{subfigure}{.33\textwidth}
\centering
\includegraphics[width=\linewidth]{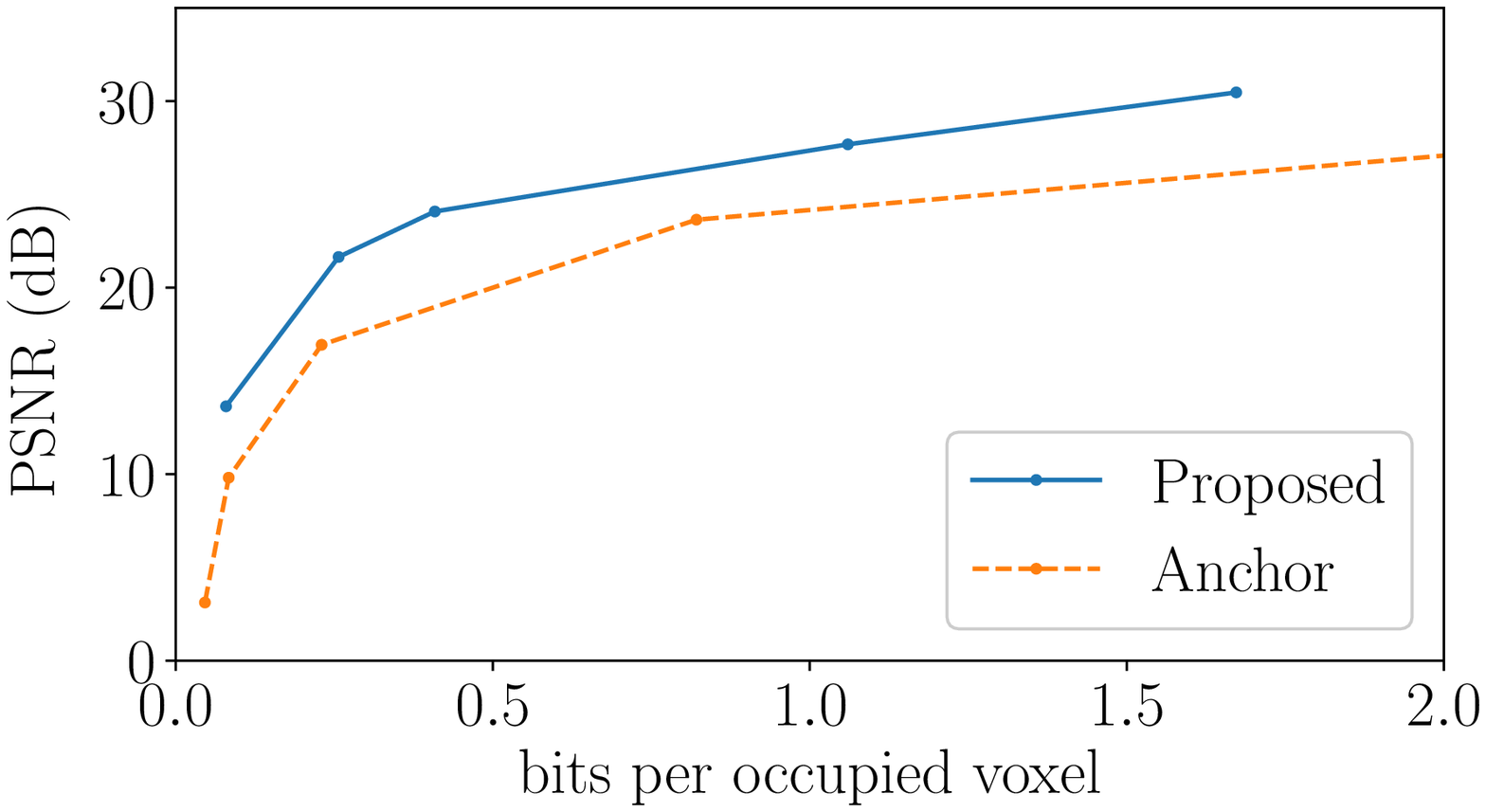}
\caption{Sarah sequence ($-52.4\%$ BDBR)}
\end{subfigure}
\hfill
\caption{RD curves for each sequence of the MVUB dataset. We compare our method to the MPEG anchor.}
\label{fig:RDcurves}
\end{figure*}

\begin{figure*}[htb]
\centering
\begin{subfigure}[t]{0.33\textwidth}
	\centering
	\includegraphics[width=\linewidth]{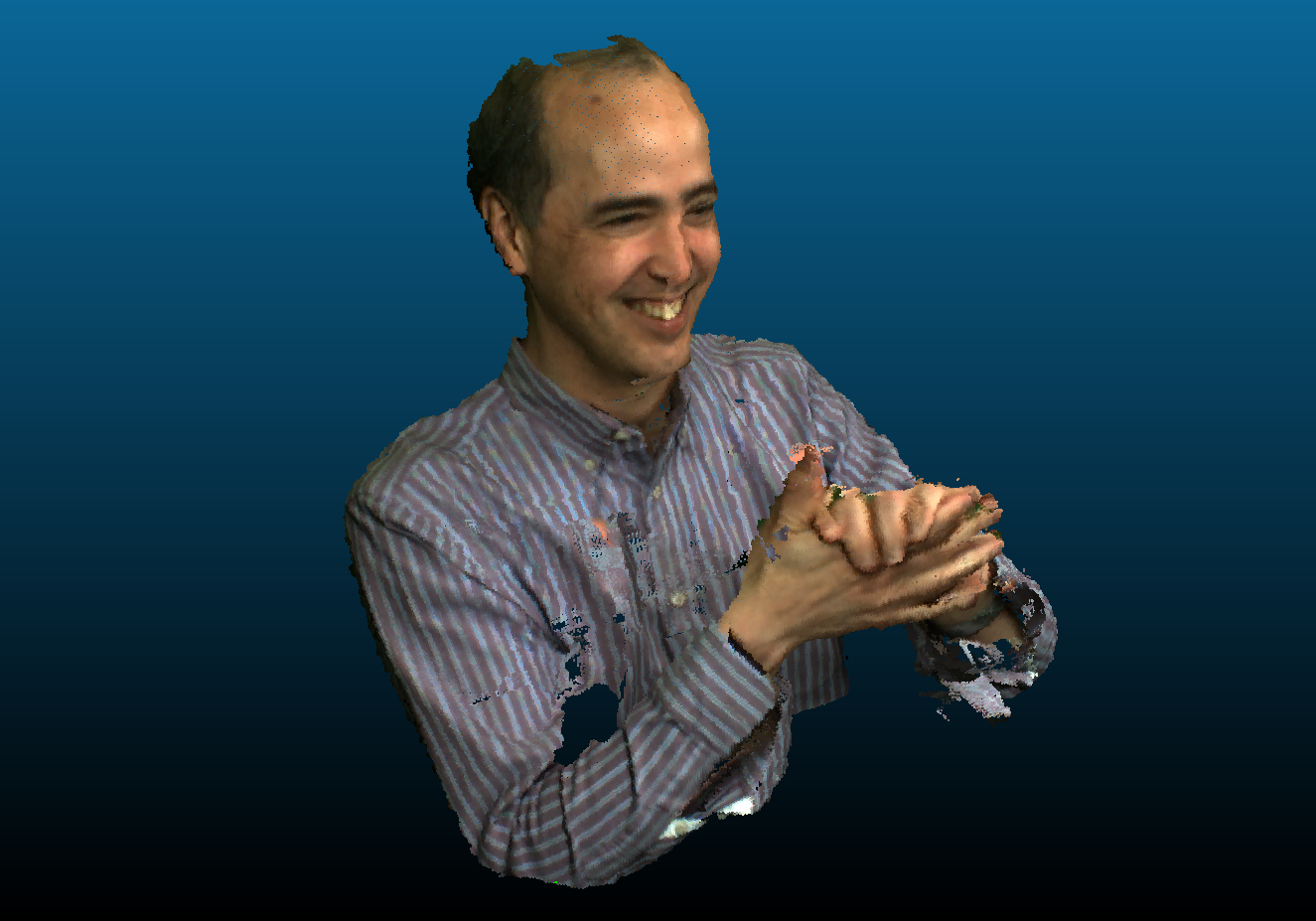}
\end{subfigure}
\hfill
\begin{subfigure}[t]{0.33\textwidth}
	\centering
	\includegraphics[width=\linewidth]{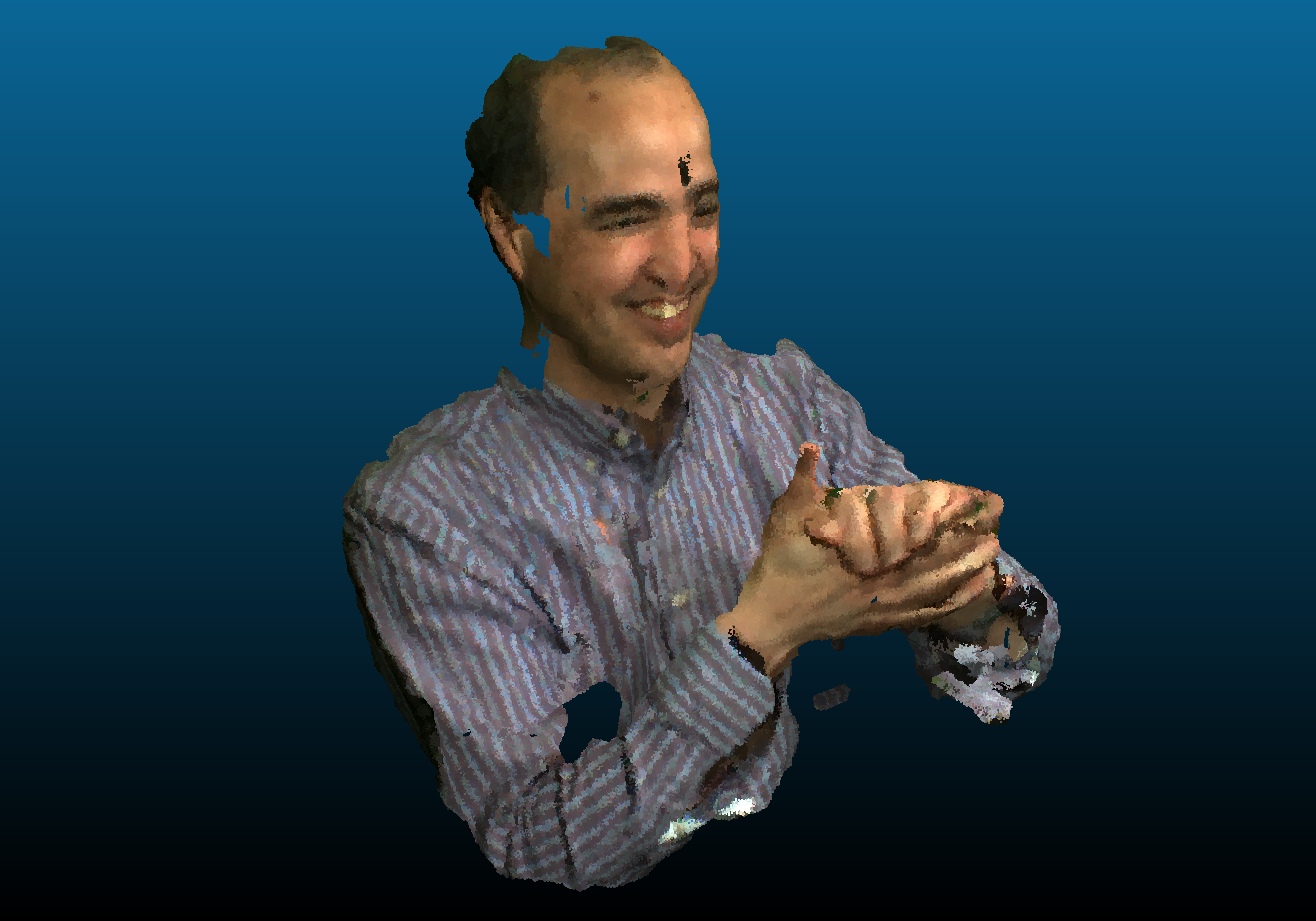}
\end{subfigure}
\hfill
\begin{subfigure}[t]{0.33\textwidth}
	\centering
	\includegraphics[width=\linewidth]{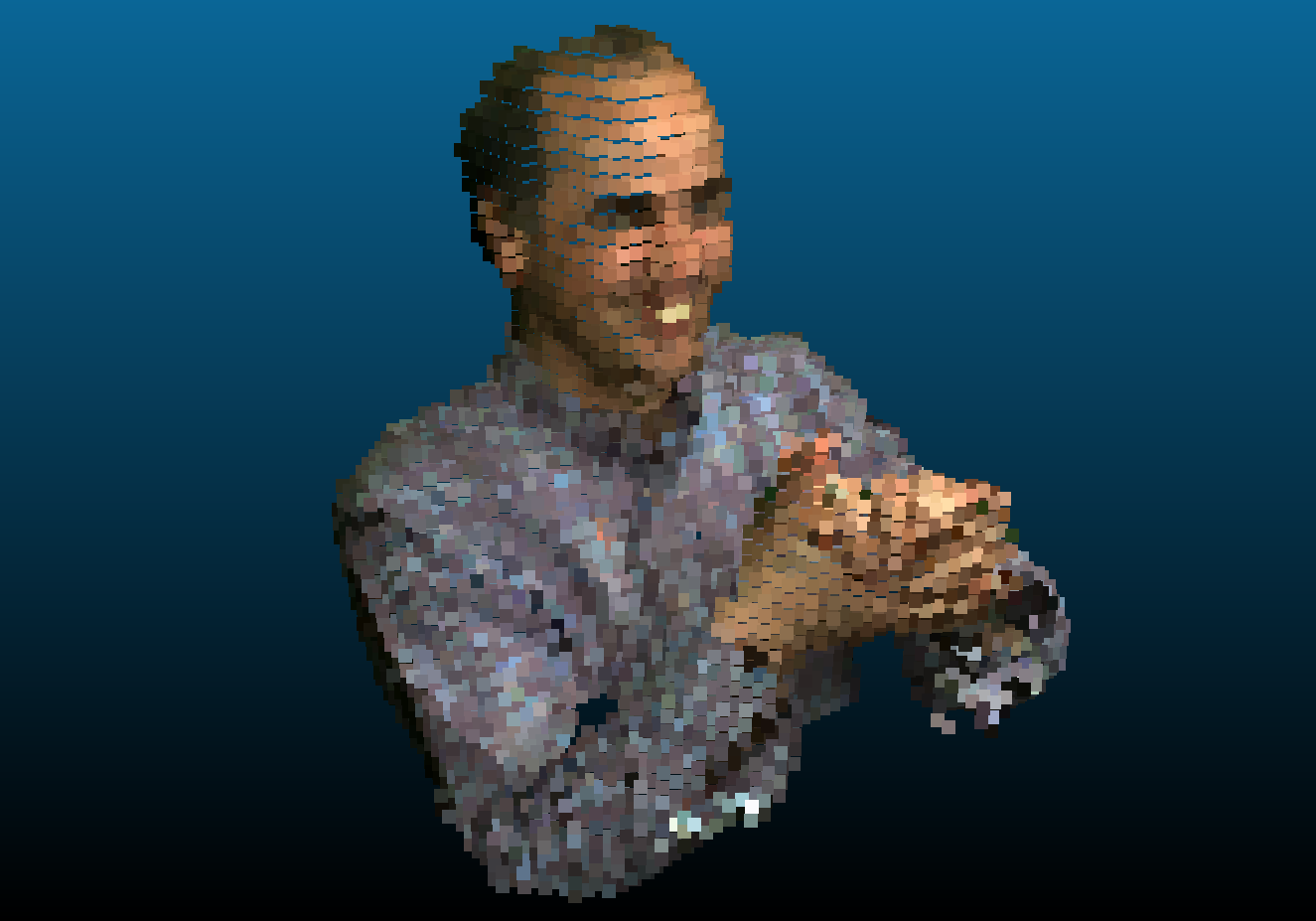}
\end{subfigure}
	\caption{
		Original point cloud (left), the compressed point cloud using the proposed method (middle) and the MPEG anchor (right).
		Colors are mapped using nearest neighbor matching.
		Our compressed point cloud was compressed using $\lambda = 10^{-6}$ with a PSNR of $29.22$ dB and $0.071$ bpov.
		The anchor compressed point cloud was compressed using a depth 6 octree with a PSNR of $23.98$ dB and $0.058$ bpov.
		They respectively have 370,798; 1,302,027; and 5,963 points.}
	\label{fig:qual}
\end{figure*}

We compute RD curves for each sequence of the test dataset.
For our method, we use the following $\lambda$ values to compute RD points : $10^{-4}$, $5 \times 10^{-5}$, $10^{-5}$, $5 \times 10^{-6}$ and $10^{-6}$.
For each sequence, we average distortions and bitrates over time for each $\lambda$ to obtain RD points.
For the MPEG anchor, we use the same process with different octree depths.

To evaluate distortion, we use the \emph{point-to-plane symmetric PSNR} \cite{tian_geometric_2017} $e_{symm}(A,B) = \min(e(A,B), e(B,A))$ where $e(A,B)$ provides the point-to-plane PSNR between points in $A$ and their nearest neighbors in $B$.
This choice is due to the fact that original and reconstructed point clouds may have a very different number of points, e.g., in octree-based methods the compressed point cloud has significantly less points than the original, while in our method it is the opposite.
In the rest of this section, we refer to the point-to-plane symmetric PSNR as simply PSNR.

Our method outperforms the MPEG anchor on all sequences at all bitrates.
The latter has a mean bitrate of $0.719$ bpov and a mean PSNR of $16.68$ dB while our method has a mean bitrate of $0.691$ and a mean PSNR of $24.11$ dB. RD curves and the Bjontegaard-delta bitrates (BDBR) for each sequence are reported in Figure~\ref{fig:RDcurves}.
Our method achieves $51.5\%$ BDBR savings on average compared to the anchor.

In Figure \ref{fig:qual}, we show examples on the first frame of the Phil sequence.
Our method achieves lower distortion at similar bitrates and produces more points than the anchor which increases quality at low bitrates while avoiding ``blocking'' effects.
This particular example shows that our method produces 218 times more points than the anchor at similar bitrates.
In other words, both methods introduce different types of distortions.
Indeed, the number of points produced by octree structures diminishes exponentially when reducing the octree depth.
Conversely, our method produces more points at lower bitrates as the focal loss penalizes false negatives more heavily.

In this work, we use a fixed threshold of $0.5$ during decompression. 
Changing this threshold can further optimize rate-distortion performance or optimize other aspects such as rendering performance (number of points).

\section{Conclusion}
\label{sec:conclusion}
We present a novel data-driven point cloud geometry compression method using learned convolutional transforms and a uniform quantizer.
Our method outperforms the MPEG Anchor on the MVUB dataset in terms of rate-distortion with $51.5\%$ BDBR savings on average.
Additionally, in constrast to octree-based methods, our model does not exhibit exponential diminution in the number of output points at lower bitrates.
This work can be extended to the compression of attributes and dynamic point clouds.

\section{Acknowledgments}
This work was funded by the ANR ReVeRy national fund (REVERY ANR-17-CE23-0020).

\bibliographystyle{IEEEbib}
\bibliography{main}

\end{document}